\documentclass{article}
\usepackage{spconf,amsmath,graphicx,hyperref}
\usepackage{amsthm,amssymb,multirow,booktabs,makecell,dsfont}
\usepackage{algorithm}
\usepackage{algpseudocode}
\usepackage{mymath}
\usepackage{cite}
\newtheoremstyle{compact_style}
  {0pt} 
  {0pt} 
  {\itshape} 
  {} 
  {\bfseries} 
  {.} 
  {.5em} 
  {} 

\theoremstyle{compact_style}

\newtheorem{theorem}{Theorem}
\newtheorem{assumption}{Assumption}
\newtheorem{remark}{Remark}
\newtheorem{lemma}{Lemma}

\title{ParaAegis: Parallel Protection for Flexible Privacy-preserved Federated Learning}
\name{Zihou Wu$^\star$, Yuecheng Li$^\star$, Tianchi Liao$^\dagger$, Jian Lou$^\dagger$, Chuan Chen$^{\star,\ast}$}
\address{$^\star$School of Computer Science and Engineering, Sun Yat-sen University, Guangzhou, China \\ $^\dagger$School of Software Engineering, Sun Yat-sen University, Zhuhai, China}
\begin{document}
%
\maketitle

\begin{abstract}
Federated learning (FL) faces a critical dilemma:
existing protection mechanisms like differential privacy (DP) and homomorphic encryption (HE) enforce a rigid trade-off,
forcing a choice between model utility and computational efficiency.
This lack of flexibility hinders the practical implementation.
To address this, we introduce ParaAegis,
a parallel protection framework designed to give practitioners flexible control over the privacy-utility-efficiency balance.
Our core innovation is a strategic model partitioning scheme.
By applying lightweight DP to the less critical, low norm portion of the model while protecting the remainder with HE,
we create a tunable system.
A distributed voting mechanism ensures consensus on this partitioning.
Theoretical analysis confirms the adjustments between efficiency and utility with the same privacy.
Crucially, the experimental results demonstrate that by adjusting the hyperparameters,
our method enables flexible prioritization between model accuracy and training time.
\end{abstract}

\section{Introduction}
\label{sec:intro}
Federated learning (FL), a distributed machine learning paradigm \cite{mcmahan2017communication},
has garnered increasing interest on account of its capability to facilitate limited data flow between data silos.
However, while FL has distinct advantages, it also faces potential challenges,
with those concerning privacy and security being a significant research direction at present.
Since this process involves transferring information over a complex network environment,
malicious devices can more easily eavesdrop on the link and intercept the transmitted models.
Although the local data is not directly included in the model parameters,
there is research that demonstrates that it is possible to reconstruct the local training data
from the transmitted models \cite{geiping2020inverting,athalye2018obfuscated}.

Differential privacy (DP) and homomorphic encryption (HE) are the most widely used to ensure privacy and security.
DP provides a formal mathematical framework to measure the privacy of a system.
In the context of FL, it typically involves techniques like gradient clipping and noise addition to meet the DP requirements.
On the other hand, HE enables computations like addition or multiplication to be performed on encrypted data, without revealing the underlying information.
DP tends to reduce model accuracy, with the impact growing as the level of privacy protection increases\cite{mcmahan2018learning}.
HE, while preserving data security, significantly increases both the training time and communication overhead\cite{ma2022privacy}.
Consequently, balancing privacy, utility, and efficiency in FL remains a significant challenge,
and current research is exploring hybrid approaches that combine different protection methods to address these trade-offs.

Some research has explored the potential of balancing privacy, accuracy and efficiency
\cite{sebert2023combining,hu2024maskcrypt,li2025clients}.
However, these works do not further investigate the optimization of the implicit trade-offs between privacy, utility, and efficiency.
To tackle this issue, we propose ParaAegis, a DP-HE-parallelly-utilizing FL framework, where the model parameters are partitioned into two parts and protected by DP and HE respectively.
To address the issue of model partition consistency accross clients, we propose a voting mechanism. In this mechanism, clients upload its local partition to the server, which then counts the occurrences of each index and selects the most frequent ones as the global partition.
Our theoretical analysis and emprical results show that the model partition for parallel protection, involving DP and HE, provides significant flexibility on privacy-utility-efficiency trade-offs. Moreover, with a fixed partition ratio, the partition strategy selecting HE part based on the maximum norm further enhances model accuracy without compromising privacy and efficiency.
The code is available on \url{https://anonymous.4open.science/r/ParaAegis/}.

\section{Preliminary}
\label{sec:pre}

In the context of FL, we consider $n$ clients, each of which has a distinct local dataset. The local objective function for the $i$-th client is denoted by $f_i := \sum_{(\mathbf{x},y)\in\mathcal{D}_i} \ell_i (\mathbf{w};\mathbf{x},y)$, where $\mathbf{w}$ and $(\mathbf{x},y)$ are the parameters and samples respectively. The global objective is the weighted average of the local objectives, weighted by the number of data samples on each client. During the learning process, the clients and the server alternately exchange local updates $\mathbf{u}_i$ and global updates $\mathbf{u}$.

To protect transmitted updates from inversion attacks, some methods propose applying the $\theta$-norm clipping and Gaussian noise injection to provide theoretical guarantees of client-level DP (see \cite{mcmahan2018learning} for the formal definition):

\begin{equation}
    \hat{\mathbf{u}}_i := \frac{\mathbf{u}_i}{\max \{1,\normt{\mathbf{u}_i}\/ \theta\}} + \mathbf{z}_i \sim \mathcal{N}(0,\sigma_z \mathbf{I}).
    \label{eq:dp}
\end{equation}
DP has a negligible impact on efficiency. However, its adverse effect on utility primarily stems from clipping and noise addition, and some studies have indicated a negative correlation between the magnitude of these operations and model accuracy.

The aforementioned methods often sacrifice model accuracy. An alternative approach, which prioritizes utility at the expense of efficiency, HE \cite{rivest1990cryptography}. With this cryptographic method, clients encrypt their local updates into ciphertexts $\cipher{\mathbf{u}_i}$ using a public key. The server then aggregates these ciphertexts in the encrypted domain by leveraging the homomorphic addition primitive, resulting in an encrypted global update  $\cipher{\mathbf{u}}$. Finally, the clients can decrypt this result using the private key. Throughout this process, the cryptographic properties of HE guarantee that all information remains computationally secure. HE has a negligible impact on utility, while its efficiency overhead typically scales linearly with the plaintext length. Among the various HE schemes, the CKKS cryptosystem \cite{cheon2017homomorphic} is particularly efficient for FL, owing to its plaintext space supporting real-valued vectors and its inherent batching capabilities. Therefore, the HE component in this paper is based on the CKKS scheme.
\section{Methodology}
\label{sec:method}

\begin{algorithm}[tb]
\caption{Proposed ParaAegis}
\label{alg:hybrid}
\algnewcommand{\Input}[1]{\item[\textbf{Input:}] #1}
\algnewcommand{\Output}[1]{\item[\textbf{Output:}] #1}
\begin{algorithmic}[1]
\State Initialize the global model $\mathbf{w}_0$ and broadcast it to clients
\For{$t = 1$ to $T$}
    \State Randomly sample clients $\mathcal{C}_t \subseteq [M]$
    \ForAll{$i \in \mathcal{C}_t$ \textbf{in parallel}}
        \State Train the model for $K$ epochs locally;
        \State Compute the local update $\mathbf{u}_i^t := \mathbf{w}_i^{t, K} - \mathbf{w}_i^{t}$;
        \State Select the indices of $\mathbf{u}_i^t$ with the highest $r\%$ norm to construct a partition vector $\mathbf{v}_i^t$;
        \State Upload $\mathbf{v}_i^t$ to the server;
    \EndFor
    \State Aggregate the partition $\mathbf{v}^t := \textsc{Voting}(\mathbf{v}_i^t)$;
    \State Send $\mathbf{v}^t$ to clients;
    \ForAll{$i \in \mathcal{C}_t$ \textbf{in parallel}}
        \State Divide $\mathbf{u}_i^t$ into $\mathbf{u}_{i,\text{DP}}^t$ and $\mathbf{u}_{i,\text{HE}}^t$ according to $\mathbf{v}^t$;
        \State Perturb $\mathbf{u}_{i,\text{DP}}^t$ to $\hat{\mathbf{u}}_{i,\text{DP}}^t$ by Eq. \ref{eq:dp};
        \State Encrypt $\mathbf{u}_{i,\text{HE}}^t$ by public key to $\cipher{\mathbf{u}_{i,\text{HE}}^t}$;
        \State Upload $\hat{\mathbf{u}}_{i,\text{DP}}^t$ and $\cipher{\mathbf{u}_{i,\text{HE}}^t}$ to the server;
    \EndFor
    \State Aggregate updates into $\hat{\mathbf{u}}_\text{DP}^t$  and $\cipher{\mathbf{u}_{\text{HE}}^t}$ seperatedly, then send them to all clients;
    \ForAll{$i = 0$ to $M-1$ \textbf{in parallel}}
        \State Decrypt $\cipher{\mathbf{u}_{\text{HE}}^t}$ to $\mathbf{u}_{\text{HE}}^t$ by secret key;
        \State Reorganize $\hat{\mathbf{u}}_\text{DP}^t$ and $\mathbf{u}_{\text{HE}}^t$ into $\mathbf{u}^t$ by $\mathbf{v}^t$;
        \State $\mathbf{w}^{t+1} := \mathbf{w}^t + \mathbf{u}^t$;
    \EndFor
\EndFor
\end{algorithmic}
\end{algorithm}

\begin{figure}[tb]
    \centering
    \includegraphics[width=0.4\textwidth]{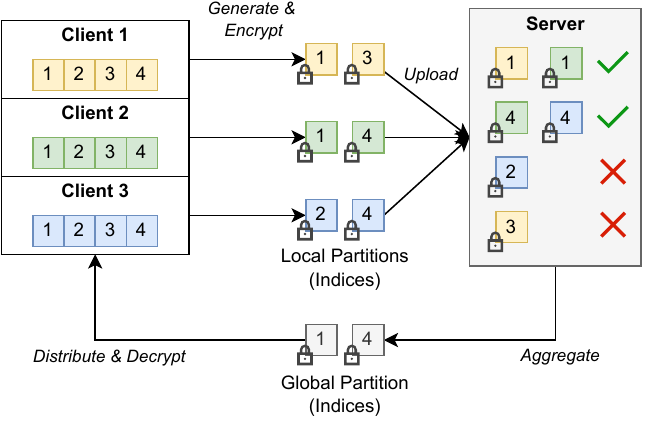}
    \caption{Illustration of the \textsc{Voting} mechanism. Each client proposes indices for HE protection. The server aggregates these proposals and selects the top-k most frequently proposed indices (e.g., indices 1 and 4) to form the global partition for all clients.}
    \label{fig:voting}
\end{figure}

In this section, we elaborate on the proposed DP-HE parallel protection method ParaAegis. Our approach is detailed in Alg.\ref{alg:hybrid} and motivated by the observation that not all parameters in a deep learning model contribute equally to the learning process. Updates with larger norms tend to be more influential in guiding the model's convergence. However, as shown in Eq.\ref{eq:dp}, these large-norm updates are also the most severely impacted by the clipping and noise injection inherent in DP, which can harm model utility.

The core tenet of our framework is to partition the local update vector, $\mathbf{u}_i^t$, into two disjoint subsets. A small, high-norm subset, which we denote $\mathbf{u}_{i,\text{HE}}^t$, is protected by HE, preserving its precision. The remaining majority of parameters, $\mathbf{u}_{i,\text{DP}}^t$, are protected by the DP mechanism. This partitioning allows the DP-protected group to use a much smaller clipping threshold, thereby reducing noise and preserving utility. Concurrently, limiting the application of HE to a small subset minimizes its computational and communication overhead.

The vector-based nature of the CKKS scheme necessitates a common partition for all clients, requiring a consensus on the partition indices. We achieve this consensus via a server-aggregated voting mechanism inspired by FL model aggregation (Fig. \ref{fig:voting}). In this process, each client "votes" for indices of its highest-norm parameters as $\mathbf{v}_i$. However, transmitting these indices in plaintext would pose a significant privacy risk by revealing information about local updates. To mitigate this, indices are encrypted. Since the server's task is limited to counting votes, standard non-homomorphic encryption is sufficient. The server receives all encrypted local partitions, determines the global partition by selecting the most frequently occurring indices (identifiable as unique ciphertexts), and then distributes this partition to all clients to apply to their respective updates.

The trade-offs inherent in the ParaAegis framework are adjustable through the partition ratio. For simplicity, we define $r$ as the proportion of the total parameters allocated to the HE-protected partition, with the remaining $(1-r)$ portion being protected by DP. This ratio acts as a crucial lever: as $r$ decreases, protection leans more on DP, enhancing efficiency at a potential cost to utility, while a larger $r$ shifts the balance towards HE, improving precision at the expense of computational overhead.

Crucially, in ParaAegis, this partition ratio is not required to be fixed, allowing for a more flexible approach tailored to the training dynamics. This is motivated by the observation that in the early stages of training, backpropagation gradients typically have larger norms and establish the general direction of convergence; interfering with them via noise can slow down the process. Conversely, gradients in later stages tend to have smaller norms and can benefit from the regularizing effect of randomness, making them more tolerant to noise. Inspired by this, we designed a dynamic decay mechanism for the partition ratio, where the number of HE-protected parameters gradually decreases as training progresses. We set an initial HE ratio $r_0$ and a decay rate $\lambda \in (0,1)$, and after each round, the proportion of HE-protected parameters is updated according to $r_{t} := \lambda r_{t-1}$. This mechanism allows us to focus on accuracy during the early training stages by minimizing deviations in the convergence path, and then shift emphasis towards efficiency in the later stages by reducing the computational cost of HE, thereby minimizing training time while preserving model utility. We validate the rationale of this design in the subsequent experimental analysis.
\section{Analysis}
\label{sec:theory}
We establish the privacy and convergence guarantees for ParaAegis.
Due to space constraints, we present the main theorems. 
The complete, detailed proofs are available in Appendix \ref{sec:app}.
\begin{assumption}
    \label{asum}
    We assume standard conditions for FL convergence analysis \cite{zhang2022understanding}: a smooth loss function $f$ with bounded gradient norms and variances across clients and data samples.
\end{assumption}

\begin{theorem}[Convergence of ParaAegis]
    \label{thm}
    Under Assumption \ref{asum}, and if the noise injected satisfies $(\varepsilon,\delta)$-DP, the convergence of ParaAegis is bounded by
    
    \begin{align}
    \frac{1}{T} & \sum_{t=0}^{T-1} \mathbb{E}[\|\nabla f(\mathbf{w}^t)\|^2] \le \\&O\left(\frac{1}{T}\right) + \underbrace{C_1 \cdot (1-r)}_{\text{Clipping Effect}} \notag + \underbrace{\frac{C_2 \cdot (1-r) \ln(1/\delta)}{N^2 \epsilon^2}}_{\text{Privacy Noise}},
    \end{align}
    where $C_1$ and $C_2$ are constants that depend on problem parameters such as the gradient bound and Lipschitz constants, and $N$ is number of clients.
\end{theorem}

By adjusting the partition ratio $r$ and privacy budget $(\varepsilon,\delta)$,
it is clear that with the privacy and efficiency changing directly, the utility,
which is represented by the convergence bound shown in Theorem \ref{thm} is affected correspondingly.
\begin{table*}[tb]
    \centering
    \small 
    \renewcommand{\arraystretch}{0.85} 
    \setlength{\tabcolsep}{4pt} 

    \begin{tabular}{@{}ll *{3}{ccc}@{}}
        \toprule
        \multirow{2}{*}{\textbf{Category}} & \multirow{2}{*}{\textbf{Algorithm}} & \multicolumn{3}{c}{\textbf{Accuracy (\%)}} & \multicolumn{3}{c}{\textbf{Time (s)}} & \multicolumn{3}{c}{\textbf{Efficiency Ratio}} \\
        \cmidrule(lr){3-5} \cmidrule(lr){6-8} \cmidrule(lr){9-11}

        & & IID & Dir(1) & Dir(0.5) & IID & Dir(1) & Dir(0.5) & IID & Dir(1) & Dir(0.5)\\
        \midrule
        
        No Privacy & FedAvg~\cite{mcmahan2017communication} & 80.93 & 79.38 & 77.98 & 3571 & 3665 & 3699 & 2.27 & 2.17 & 2.11 \\
        \midrule
        \multirow{5}{*}{Baselines} & DP-FedAvg~\cite{wei2020federated} & 20.28 & 17.12 & 12.87 & 3007 & 3084 & 3112 & 0.67 & 0.56 & 0.41 \\
        & CKKS-FedAvg~\cite{qiu2022privacy} & 81.14 & 79.28 & 77.63 & 18527 & 14971 & 14568 & 0.44 & 0.53 & 0.53 \\
        & Serial~\cite{zhang2024secure} & 10.54 & 11.29 & 12.35 & 19956 & 16718 & 16370 & 0.05 & 0.07 & 0.08 \\
        & Varying DP~\cite{yuan2023amplitude} & 24.21 & 21.76 & 19.85 & 2494 & 2534 & 2789 & 0.97 & 0.86 & 0.71 \\
        & Stevens et al.~\cite{stevens2022efficient} & 18.78 & 17.07 & 13.75 & 3635 & 3708 & 3543 & 0.52 & 0.46 & 0.39 \\
        \midrule
        \multirow{4}{*}{ParaAegis-Static}
        & \(1\%\)  & 67.85 & 63.48 & 57.16 & 6149 & 6006 & 5577 & 1.1 & 1.06 & 1.02 \\
        & \(5\%\)  & 73.03 & 70.07 & 66.19 & 6888 & 6962 & 6099 & 1.06 & 1.01 & 1.09 \\
        & \(10\%\) & 75.63 & 73.63 & 71.1 & 7776 & 7706 & 6379 & 0.97 & 0.96 & 1.11 \\
        & \(20\%\) & 77.53 & 75.37 & 73.39 & 9530 & 9565 & 7153 & 0.81 & 0.79 & 1.03 \\
        \midrule
        \multirow{4}{*}{ParaAegis-Dynamic}
        & \(5\% \times 0.99\) & 72.09 & 68.91 & 64.4 & 5561 & 5536 & 5299 & \textbf{1.3} & \textbf{1.24} & \textbf{1.22} \\
        & \(10\% \times 0.99\) & 74.5 & 71.75 & 68.5 & 6238 & 6210 & 5679 & 1.19 & \underline{1.16} & \underline{1.21} \\
        & \(10\% \times 0.95\) & 68.82 & 65.13 & 58.97 & 5619 & 5628 & 5400 & \underline{1.22} & \underline{1.16} & 1.09 \\
        & \(10\% \times 0.9\) & 70.87 & 67.01 & 60.33 & 8762 & 8732 & 7073 & 0.81 & 0.77 & 0.85 \\
        \bottomrule
    \end{tabular}
    \caption{Comparison on Performance across Different Algorithms With ResNet-18. The Efficiency Ratio is calculated as (Accuracy / Time) $\times 100$. For ParaAegis-Static, the parameter is the fixed ratio \( r \). For ParaAegis-Dynamic, it is \( r_0 \times \lambda \).}
    \label{tab:perf-resnet-restructured}
\end{table*}

\section{Experiments}
\label{sec:experiment}

\textbf{Experimental Setup.}
We conduct experiments on Imagenette \cite{imagenette2019} dataset and ResNet-18 model~\cite{he2016deep},
whose parameter sizes are sufficient to highlight the advantages of our method in balancing privacy, efficiency, and utility. For the federated learning setup, the total number of clients is $N=10$. The training proceeds for a total of $T=50$ global rounds, with each client performing $K=3$ local epochs per round. Regarding the training hyperparameters, the batch size $|\mathcal{B}|$ is set to 32, the learning rate $\eta$ is 0.01, and the gradient clipping threshold $\theta$ is 1.

\textbf{Benchmarks.} To evaluate the flexibility of Paraaegis with respect to the privacy-utility-efficiency trade-off, we benchmark it against CKKS-FedAvg~\cite{qiu2022privacy} and DP-FedAvg~\cite{wei2020federated}.
We also consider the serial combination of DP and HE~\cite{zhang2024secure} as a competitive alternative.
Additionally, two methods considering utility-efficiency trade-off with privacy guarantee are included,
where the former utilizes amplitude-varying DP~\cite{yuan2023amplitude} with the scale factor set as 0.9,
and the latter employs DP, secret sharing and learn-with-errors scheme~\cite{stevens2022efficient}.
To monitor an environment with relatively strict privacy constrains, the privacy budget of DP is set to \((\varepsilon,\delta)=(1,10^{-5})\) 
Table \ref{tab:perf-resnet-restructured} records the comparison of classification accuracy and running time between Paraaegis and the benchmark methods under various parameter settings.
For a more intuitive presentation of the trade-off between performance and efficiency,
we define the ratio of accuracy to training time as the \textit{Efficiency Ratio},
which reflects the quality of this trade-off. Here are the findings observed from Table \ref{tab:perf-resnet-restructured}.
(1) As the HE proportion $r$ increases,
the model accuracy improves accordingly, but the time overhead also increases.
(2) DP-FedAvg and CKKS-FedAvg can be considered as extreme cases of ParaAegis.
In their trade-offs between utility and efficiency,
each method sacrifices one aspect to gain an advantage in the other.
This also demonstrates that ParaAegis's parallel combination of DP and HE
provides more flexibility in adjusting the efficiency-utility trade-off for model training.
(3) The serial combination of the two protection methods results
in a simultaneous reduction in both accuracy and efficiency,
making it unsuitable for scenarios that require a balance between efficiency and utility.
(4) Two extra methods (Varying DP and Steven et al.) perform relatively worse due to their complete reliance on differential privacy, whereas ParaAegis can avoid this drawback.
(5) In the dynamic variant, selecting appropriate parameters $r$ and  $\lambda$ can
achieve better accuracy in less time compared to the static version,
resulting in an improved privacy-efficiency-utility trade-off.

\begin{table}[tb]
    \centering

    \begin{tabular}{lccc}
        \toprule
        Strategy & Max (Proposed) & Min & Rand \\
        \midrule
        Accuracy (\%) & 63.64 & 12.40  & 15.78 \\
        \bottomrule
    \end{tabular}
    \caption{Comparison on Accuracy across Partition Strategies}
    \label{tab:strategies}
\end{table}

\begin{table}[tb]
    \small
    \centering
    \begin{tabular}{l|cccc}
        \toprule
        $\theta$ & 0.01 & 0.1 & 1 & 10 \\
        Acc. & 68.99 & 69.11 & 71.88 & 65.16 \\
        \midrule
        $N$ & 5 & 10 & 25 & 50 \\
        Acc. & 72.49 & 65.16 & 56.40 & 48.14 \\
        \bottomrule
    \end{tabular}
    \caption{Comparison on Accuracy across Clipping Thresholds or \# of Clients}
    \label{clipping}
\end{table}

\textbf{Ablation of Partition Strategies.}
In this experiment, we fix the partition ratios and compare them across different partition strategies to demonstrate the advantages of the partition strategy our proposed.
The partitioning strategies involved in the experiment are as follows.
\textit{Max}, as the strategy adopted in ParaAegis, selects the parameter set with the highest norm from local updates of each client as the HE part; \textit{Min}, in contrast to \textit{Max}, selects the parameter set with the lowest norm from the local updates of each client as the HE part; \textit{Rand}, the client randomly selects a certain number of parameters as the DP part. To facilitate observing the differences, we fixed the partition ratio at $r = 0.1$ and number of global rounds at $T = 20$. Observed from Table \ref{tab:strategies} we know that the \textit{Min} and \textit{Rand} strategies exhibit similarly poor and unacceptable convergence behaviors,
while only \textit{Max} achieves well-behaved accuracy.
This phenomenon indicates that the \textit{Max} strategy
effectively selects the most important parameters for HE protection,
shielding them from the perturbations introduced by noise.

\textbf{Hyperparameter Sensitivities.}
We evaluate ParaAegis's sensitivity to two key hyperparameters: the DP clipping threshold $\theta$ and the number of clients $N$. First, we determined the optimal clipping threshold by testing $\theta$. The results in Table \ref{clipping} indicate that $\theta=1$ yields the best accuracy. This optimal value is significantly smaller than in typical DP-FL settings \cite{ponomareva2023dp}, which demonstrates that our voting mechanism effectively reduces the norm of the update component subjected to DP noise. Second, we examined the impact of client scale by varying $N$. The observed decrease in accuracy with a larger $N$ (Table \ref{clipping}) indicates that reaching a consensus becomes more difficult with increased client diversity. Consequently, ParaAegis is best suited for cross-silo FL environments, a finding consistent with other HE-based FL frameworks \cite{blanco2021achieving}.
\section{Conclusion}
\label{sec:end}

In this paper, we propose a novel parallel hybrid protective method for FL.
Our method partition the model parameters into two part,
both of which are protected by DP and HE respectively,
and is capable to effectively trade off privacy, utility and efficiency flexibly
via adjusting partition ratio $r$ and decay rate $\lambda$ (in dynamic variant).

\vfill\pagebreak

\bibliographystyle{IEEEbib}
\bibliography{references}
\newpage
\appendix
\section{Detailed Theorem Analysis}
\label{sec:app}
In this section, we provide detailed mathmatical expression of privacy-utility-efficiency trade-off.

\subsection{Privacy Analysis}

On account for the Gaussian noise mechanism in proposed method,
the privacy budget can be calculated by moment accountant theory\cite{abadi2016deep},
which is mainly Lem. \ref{lem:abadi}.

\begin{lemma}
    \label{lem:abadi}
    There exists constant $c_1, c_2$ such that for any $\varepsilon \leq c_1 q^2 T$,
    where $q = \frac{n}{N}$, FedAvg is $(\varepsilon, \delta)$-DP for any $\delta > 0$ if we choose
    \[\sigma_z \geq c_2 \frac{n^2T \ln (1/\delta)}{N^2 \varepsilon^2}.\]
\end{lemma}

\begin{remark} \label{rem:1}
    Moreover, noise amplitude can be expressed by privacy budget\cite{yuan2023amplitude, wei2020federated}, i.e.,
    \[\sigma_z^2 = (\frac{\Delta f}{\varepsilon})^2 \cdot 2qT \ln (1/\delta),\]
    where $\Delta f$ is the sensitivity of the object function.
    In the context of this paper,
    sensitivity refers to the maximum difference in the DP part norm
    between two consecutive updates uploaded by a client.
    According to related researchs, sensitivity can be defined as
    \[\Delta f = \min_i \frac{2 \theta}{\left|\mathcal{D}_i\right|}.\]
\end{remark}

The aforementioned privacy budget refers solely to the DP part of the local update,
and concerning HE part we can utilized the properties of cryptographic algorithms.
Due to computational security of current public key cryptosystems \cite{rivest1990cryptography},
the probability that an adversary without the key can obtain any information,
including membership information related to the definition of DP,
from the encrypted update is negligible,
as long as the key length is sufficiently long.
Thus, the privacy of ParaAegis relies entirely on the DP part,
which is similar to the barrel effect. 

We conducts the proof of Theorem 1.

\begin{proof}
    Due to the cryptographic properties of homomorphic encryption, the amount of information that can be extracted from the HE part is computationally negligible, and thus the privacy budget consumed by HE protection is also nearly zero. In parallel, according to Lemma 1, we establish the relationship between the noise magnitude in the DP part and the corresponding privacy budget. Furthermore, by the parallel composition theorem of DP \cite{dwork2014algorithmic}, the overall privacy budget for gradient exposure is determined solely by the privacy budget allocated to the DP part.
\end{proof}

\subsection{Detailed Assumptions}

The following assumptions is utilized in the convergence of the FL problem under proposed hybrid protection, which are detailed version of Assumption \ref{asum}.

\begin{assumption}[Lipschitz Smooth]
    \label{asm:lip}
    There exists $L$ such that for any $\bd{x}, \bd{y}, i$ it holds that \(\norm{\nabla f_i (\bd{x}) - \nabla f_i (\bd{y})} \leq L \norm{\bd{x} - \bd{y}}\).
\end{assumption}

\begin{assumption}[Bounded Local Variance]
    There exists $\sigma_l$ such that for any $t, k, i$ it holds that \(\normt{\gitk - \nabla f_i(w_i^{t,k})} \leq \sigma_l^2\).
\end{assumption}

\begin{assumption}[Bounded Global Variance]
    There exists $\sigma_g$ such that for any $i$ it holds that \(\nabla f_i(\bd{w}) - \nabla f(\bd{w}) \leq \sigma_g^2\).
\end{assumption}

\begin{assumption}[Bounded Gradient]
    \label{asm:bg}
    There exists $G$ such that for any $t,k,i$ it holds that \(\normt{\gitk} \leq G^2\).
\end{assumption}

\subsection{Proof for Theorem 1}

In this subsection, we outline the necessary preparation work required
before delving into the detailed proof of convergence.

Firstly, we clarify the definitions of the DP part and the HE part,
i.e., for any vector \(\bd{x}\) and any \(l \in [\operatorname{dim} \bd{x}]\),
its HE part \(\bd{x}_{\text{HE}}\) and DP part \(\bd{x}_{\text{DP}}\) are defined as

\begin{align*}
    \bd{x}_{\text{HE}}[l] = \bd{x}[l] \cdot \mathds{1} \{l \in \bd{v}\} \\
    \bd{x}_{\text{DP}}[l] = \bd{x}[l] \cdot \mathds{1} \{l \notin \bd{v}\}
\end{align*}

We list the clipping coefficient, its mean, and the difference, found by Zhang et al. \cite{zhang2022understanding}.

\begin{align*}
    \ait &:= \min \left(1, \frac{\theta}{\sum_{k=0}^{K-1} \norm{\guitkdp}}\right), \\
    \bat &:= \frac{1}{N} \sum_{i=1}^{N} \ait, \ \tat := \frac{1}{N} \sum_{i=1}^{N} \left| \ait - \bat \right|.
\end{align*}

The local update can be easily represented in the following form based on the above definition.

\begin{equation*}
    \uit := \bd{w}^{t+1} - \bd{w}^t = - \eta \sum_{k=0}^{K-1} \left( \ait \guitkdp +  \guitkhe \right)
\end{equation*}

By Lipschitz smoothness, we have
{
\small
\begin{align*}
    &\expc{f (\bd{w}^{t+1})} - \expc{\fw{t}} \\
    \leq& \expc{\inner{\nabla f(\bd{w}^t), \bd{w}^{t+1} - \bd{w}^t}} + \frac{L}{2} \expc {\normt{\ww{t+1} - \ww{t}}}\\
    =& \expc{\inner{\nabla \fw{t}, \frac{1}{n} \sum_{i \in \client} \uit}} + \frac{L}{2} \expc{\normt{\frac{1}{n} \sum_{i \in \client} \left( \uit + \bd{z}_i \right)}} \\
    =& \underbrace{\inner{\nabla \fw{t}, \expc{\frac{1}{n} \sum_{i \in \client} \uit}}}_{A_1} + \underbrace{\frac{L}{2} \expc{\normt{\frac{1}{n} \sum_{i \in \client} \uit}}}_{A_2}+ \underbrace{\frac{L \sigma_z^2 \theta^2 r d}{2 n^2}}_\text{caused by noise} \tag{*}\label{eq:efw}
\end{align*}
}

where $\bd{z}_i = [\mathds{1}\{l \in \bd{v}_i\} \cdot \mathcal{N}(0, \sigma_z)]_l$ and last term is caused by noise.

For $A_1$, we transform the expectation over sampled clients into an expectation over all clients.
 
\begin{align*}
    A_1 =& \inner{\nabla \fw{t}, \expc{\frac{1}{n} \sum_{i \in \client} \uit}} \\
    =& \inner{\nabla \fw{t}, \expc{\frac{1}{N} \sum_{i=1}^{N} \uit}} \\
    =& \inner{\nabla \fw{t}, \expc{\frac{1}{N} \sum_{i=1}^{N} (\uit - \buit)}} \\
    &+ \inner{\nabla \fw{t}, \expc{\frac{1}{N} \sum_{i=1}^{N} \buit}} \tag{**}\label{eq:a1}
\end{align*}

Then, we focus on the first term of (\ref{eq:a1}) and get

\begin{align*}
    & \inner{\nabla \fw{t}, \expc{\frac{1}{N} \sum_{i=1}^{N} (\uit - \buit)}} \\
    \leqa{i}& \inner{\nabla \fw{t}, \expc{\frac{1}{N} \sum_{i=1}^{N} \sum_{k=0}^{K-1} \eta |\ait - \bat| \gitkdp}} \\
    =& \frac{1}{N} \sum_{i=1}^{N} \sum_{k=0}^{K-1} \eta \expc{|\ait - \bat| \inner{\fw{t},\gitkdp}} \\
    \leqa{ii}& \eta \expc{\tat} K G^2 \bar{\rho}_t^2
\end{align*}

where (i) is derived from the definition of $\uit$ and $\buit$,
and (ii) is conducted by the Bounded Gradient Assumption.

Bounding the second term of (\ref{eq:a1}), we have

\begin{align*}
    & \inner{\nabla \fw{t}, \expc{\frac{1}{N} \sum_{i=1}^{N} \buit}} \\
    =& \expc{\inner{\nabla \fw{t}, \frac{1}{N} \sum_{i=1}^{N} \uuit}} \\
    =& \frac{-\eta \bt K}{2} \normt{\nabla\fw{t}} - \frac{\eta \bt}{2K} \expc{\normt{\frac{1}{\eta\bt N} \sum_{i=1}^{N} \uuit}} \\
    &+ \frac{\eta \bt}{2} \underbrace{\expc{\normt{\sqrt{K} \nabla\fw{t} - \frac{1}{\eta\bt N \sqrt{K}} \sum_{i=1}^{N} \uuit}}}_{A_3}
\end{align*}

where the second equation comes from that $\inner{\bd{x},\bd{y}} = -\frac{1}{2}\normt{\bd{x}} -\frac{1}{2}\normt{\bd{y}} +\frac{1}{2}\normt{\bd{x-y}}$
holds for any vector $\bd{x,y}$.
Noting $\guit := \nabla f_i (\bd{w}_i^{t})$ and $\guitk := \nabla f_i (\bd{w}_i^{t,k})$,
we decompose $A_3$ subsequently:

\begin{align*}
    A_3 =& K \expc{\normt{\nabla \fw{t} - \frac{1}{K N} \sum_{i=1}^{N} \sum_{k=0}^{K-1} (\ait \guitkdp + \guitkhe) }} \\
    =& K \expc{\normt{\frac{1}{KN}\sum_{i=1}^{N}\sum_{k=0}^{K-1} (\guit - \ait \guitkdp - \guitkhe)}} \\
    \leq& \frac{1}{N} \sum_{i=1}^{N}\sum_{k=0}^{K-1} \expc{\normt{\guit - \ait \guitkdp - \guitkhe}} \\
    \leq& \frac{1}{N} \sum_{i=1}^{N}\sum_{k=0}^{K-1} \expc{\normt{\guit - \guitk}}\\ &+ \frac{1}{N} \sum_{i=1}^{N}\sum_{k=0}^{K-1} \expc{\underbrace{\normt{\guitk - \ait\guitkdp - \guitkhe}}_{A_4}} \\
    \leq& \frac{1}{N} \sum_{i=1}^{N}\sum_{k=0}^{K-1} \left(L^2 \expc{\normt{\bd{w}^t - \bd{w}_{i}^{t,k}}} + G^2\right) \\
    \leq& L^2 5K^2 \eta^2 (\sigma_l^2 + 6 K \sigma_g^2) + L^2 30 Q^3 \eta^2 \normt{\nabla \fw{t}} + K G
\end{align*}

where the first inequality is derived from Jensen's inequality,
the third is derived from $L$-smoothness of $f$,
and the last one is conducted from Lemma 3 of \cite{reddi2021adaptive}, i.e., it holds for any $k$ that
\(\frac{1}{N} \expc{\normt{\bd{w}^t - \bd{w}_{i}^{t,k}}} \leq 5K^2 \eta^2 (\sigma_l^2 + 6 K \sigma_g^2) + 30 Q^3 \eta^2 \normt{\nabla \fw{t}}\).

Additionally, the detail of finding a upper bound of $A_4$ can be described as

\begin{align*}
    A_4 =& \normt{\guitkdp - \ait\guitkdp} = (1 - \ait)^2 \norm{\guitkdp}^2 \\
    \leq& (1-\ait)^2(1-\rho_t^2) G^2 \leq G^2
\end{align*}

where the first equality comes from the definition that $\guitkdp + \guitkhe = \guitk$
and the second inequality is derived from the fact that $\ait \leq 1, \rho_t \leq 1$.

Now we turn our attention to $A_2$ in \ref{eq:efw}.

\begin{align*}
    &\expc{\normt{\frac{1}{n} \sum_{i \in \client} \uit}} \\
    =&\expc{\normt{\frac{1}{n} \sum_{i \in \client} \sum_{k=0}^{K-1} \eta (\ait \gitkdp + \gitkhe)}} \\
    =&\expc{\frac{1}{n} \sum_{i \in \client} \sum_{k=0}^{K-1} \eta^2 \beta^2 \normt{\gitk}} \\
    \leq& \frac{\eta^2 G^2}{n^2} \expc{\sum_{i \in \client} \sum_{k=0}^{K-1} (\beta_i^{t,k})^2} = \frac{\eta^2 G^2 \expc{\bar{\beta}_t^2}}{n}
\end{align*}

Sum up the equations, we get

\begin{align*}
    \expc{\fw{t+1}} &\leq \fw{t} + \eta \tat (1 - \trt^2) K G^2 \\
    -& \frac{\eta \bbt K}{2} \normt{\nabla\fw{t}} - \frac{\eta\bbt}{2K} \expc{\normt{\frac{1}{\eta N \bbt} \sum_{i=1}^{N} \uuit}} \\
    +& \frac{\eta \bbt}{2} \left( 5 L^2 K^2 \eta^2 (\sigma_l^2 + 6 K \sigma_g^2) + 30 L^2 K^3 \eta^2 \normt{\nabla\fw{t}} + KG\right) \\
    +& \frac{\eta^2 L G^2 K}{2n} + \frac{L \sigma_z^2 \theta^2 d (1-r)}{2n^2}
\end{align*}

Simplify it by setting $\eta \leq \frac{1}{\sqrt{60} KL}$:

\begin{align*}
    \fw{t+1}
    \leq& \fw{t} - \frac{\eta \bbt K}{4} \gradt + \tat (1 - \trt^2) K G^2 \\
    &+\frac{5}{2} \eta^3 \bbt L^2 K^2 (\sigma_l^2 + 6 K \sigma_g^2) + \frac{1}{2}  \eta_l^2 \bbt K G^2 \\
    &+ \frac{\eta^2 L G^2 K}{2n} + \frac{L \sigma_z^2 \theta^2 d (1-r)}{2n^2}
\end{align*}

After taking the sum from $t = 0$ to $T-1$, dividing both side by $\frac{\eta K T}{4}$,
substuting $\sigma_z$ with $\varepsilon$ and $\delta$ by Lemma \ref{lem:abadi},
and rearranging, we obtain that

\begin{align*}
        & \frac{1}{T}\sum_{t=0}^{T-1}\expc{\bbt \gradt} \leq \frac{4}{\eta K T} \expc{\fw{0} - \fw{T}} \\
        & + 4G^2 \EEt{\tat(1 - \brt)} + 10 \eta^2 L^2 K (\sigma_l^2 + 6k \sigma_g^2) \EEt{\bbt} \\
        & + 2 G^2 \EEt{\bbt} + \frac{\eta^2 L G^2 K}{2n} + \frac{2 T L  \theta^2 d (1-r) \ln(1/\delta)}{N^2 D_\text{min}^2 \varepsilon^2 }.
\end{align*}

By omitting the terms irrelavent to $r$ and $(\varepsilon,\delta)$, we get Theorem \ref{thm}.

\end{document}